\documentclass{IOS-Book-Article}

\usepackage{mathptmx}
\usepackage{amsmath,graphicx}
\usepackage{amsfonts}
\usepackage{multirow}
\usepackage{soul}\setuldepth{article}
 
\def\hb{\hbox to 10.7 cm{}}

\begin{document}

\pagestyle{headings}
\def\thepage{}

\begin{frontmatter}              

\title{A Deep Local and Global Scene-Graph Matching for Image-Text Retrieval}

\markboth{}{April 2021\hb}

\author[A]{\fnms{Manh-Duy} \snm{Nguyen}%
\thanks{Corresponding Author: Manh-Duy Nguyen, School of Computing, Dublin City University, Dublin, Ireland; E-mail:manh.nguyen5@mail.dcu.ie}},
\author[B,C,D]{\fnms{Binh T.} \snm{Nguyen}}
and
\author[A]{\fnms{Cathal} \snm{Gurrin}}

\runningauthor{Manh-Duy et al.}
\address[A]{School of Computing, Dublin, Ireland}
\address[B]{AISIA Research Lab}
\address[C]{University of Science, Ho Chi Minh City, Vietnam}
\address[D]{Vietnam National University Ho Chi Minh City, Vietnam}



\begin{abstract}
Conventional approaches to image-text retrieval mainly focus on indexing visual objects appearing in pictures but ignore the interactions between these objects. Such objects occurrences and interactions are equivalently useful and important in this field as they are usually mentioned in the text. Scene graph presentation is a suitable method for the image-text matching challenge and obtained good results due to its ability to capture the inter-relationship information. Both images and text are represented in scene graph levels and formulate the retrieval challenge as a scene graph matching challenge. In this paper, we introduce the Local and Global Scene Graph Matching (LGSGM) model that enhances the state-of-the-art method by integrating an extra graph convolution network to capture the general information of a graph. Specifically, for a pair of scene graphs of an image and its caption, two separate models are used to learn the features of each graph's nodes and edges. Then a Siamese-structure graph convolution model is employed to embed graphs into vector forms. We finally combine the graph-level and the vector-level to calculate the similarity of this image-text pair. The empirical experiments show that our enhancement with the combination of levels can improve the performance of the baseline method by increasing the recall by more than 10$\%$ on the Flickr30k dataset.
\end{abstract}

\end{frontmatter}

\section{Introduction}
Computer vision and natural language processing are two of the most popular domains of deep learning, each of which has a wide range of applications \cite{duy2018accurate,Shih3DP20,johnson2017google}. A fusion of these two areas has raised many fascinating challenges and focused the attention of researchers. One of those topics is image-text retrieval, in which a text (image) query is provided to retrieve relevant images (texts) from a given dataset. This issue's critical point is the clear semantic gap between images and text, presenting a challenging multi-modal data retrieval challenge.

\begin{figure}[ht!]
  \centering
  \includegraphics[width=0.63\textwidth]{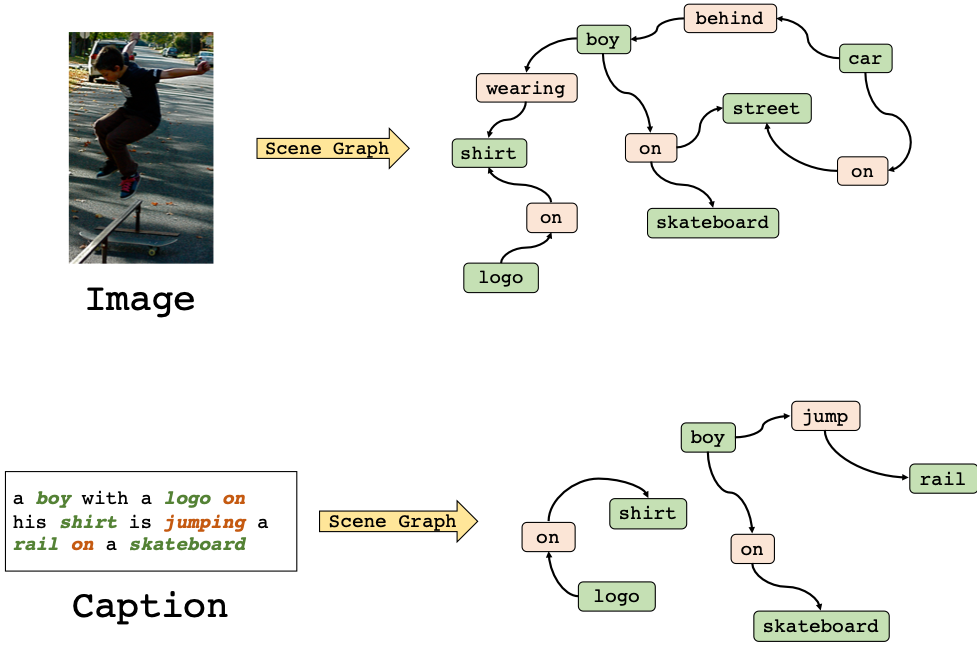}
  \caption{An example of scene graphs generated from an image and its caption. The green rectangles are nodes in the graph indicating detected objects in the image or the caption while the orange rectangles are edges illustrating the relationships between objects. }
  \label{fig:scenegraph}
\end{figure}

Many kinds of research have focused on this challenge, and one can summarise earlier works into two types. The first type includes methods \cite{faghri2017vse++,wang2017adversarial,zheng2020dualpath} in which each image and textual description are entirely encoded as global representative vectors in the same space. It can be considered a decent solution due to its simplicity and fast retrieval. However, these approaches are only suitable for simple cases where there is only a single object in an image or a short sentence as they neglect the importance of positioning information. The proposed techniques \cite{huang2018learning,lee2018stacked,wang2019position} of the second type can deal with this positioning issue. They encode images into several regions based on detected objects. Sentences are also parsed into many fields based on their chunks of words. This approach now can gain more elaborate detail for both data. Although these methods have been shown to achieve better performance than others in the first group, there are still challenges to be solved. Neither type has considered the actions occurring within the data, including the interactions between entities/objects in images and captions. Such information can offer more in-depth insights into images as well as textual queries \cite{johnson2015image}, therefore facilitating enhanced image-text retrieval. It should be mentioned that the global embedding methods in the first group also could learn this association information. However, it cannot be in detail since they do not focus directly on the relationship aspect. The introduction of a scene graph structure \cite{johnson2015image} introduces a promising way to address this issue. Due to its capacity to capture both objects and their interactions, the design has been applied and achieved excellent results in various fields \cite{johnson2018image,xu2019scene,anderson2016spice,mmm2021graph}. A scene graph structure is a graph including nodes and edges connecting nodes together. In the image-text retrieval field, a node and an edge represent objects and the associations among objects detected in images or captions, as depicted in Figure \ref{fig:scenegraph}. Recent research \cite{wang2020cross, shi2019knowledge} could obtain better results in this retrieval area by utilizing these relations information. Hence there is a promise when applying this useful data structure in the retrieval field.

In this paper, we employ a state-of-the-art method that applies scene graph structure to solve the retrieval challenge \cite{wang2020cross}. However, we advance this baseline by introducing an addition graph convolution network to capture a global form of a graph beside its local representation produced by the baseline. To be more specific, we firstly generate scene graph information of each image and its caption. Our proposed model can extract and learn the insightful features of nodes and edges from both graphs to obtain the local details. These scene graphs are then embedded into vectors by a Siamese graph model to represent the entire image's global information and textual content. The similarity of the two graphs now can be measured at the local and global levels. We evaluate our proposed method by measuring the recall metrics on the well-known Flickr30k \cite{flickr30k} dataset to show the effectiveness of our model.

\section{Related Work}
Many models have been introduced for the image-text retrieval problem. Some of them \cite{faghri2017vse++,wang2017adversarial,zheng2020dualpath} encode entire images and sentences into one space to facilitate the comparison, while other solutions focused on local information instead \cite{huang2018learning,lee2018stacked,wang2019position}. Faghri et al. \cite{faghri2017vse++} followed a typical method to use a CNN-based image encoder to extract visual features of images and an RNN-based textual module to extract those from captions. These features then went through a fully connected layer to be projected into the same vector space to facilitate the comparison. Although it is a simple approach, it still achieves high performance by applying hard negatives triplet loss during training. With different points of view on analyzing sentences, Wang et al. \cite{wang2017adversarial} used a Bag-of-Word technique to extract the information from semantic data. However, one could also use a CNN network to process sentences in a dual-path convolutional model \cite{zheng2020dualpath} by entirely embedding each image and sentence into a vector form using two CNN modules separately. In contrast, Huang et al. \cite{huang2018learning} focused on the local information in images where semantic concepts and their ordering were extracted by learning with their corresponding sentences. The work from Wang \cite{wang2019position} raised the awareness of the relative position of detected objects within an image which could be useful when matching with the caption. Meanwhile, Lee et al. \cite{lee2018stacked} presented a novel cross attention mechanism that learned the importance of regions (images) and words (captions) along with their alignment.

Although achieving considerable accuracy, previous work ignored or underestimated the robust information of objects' interactions within images and sentences. Johnson and colleagues firstly introduced the scene graph for dealing with this issue for the image retrieval challenge \cite{johnson2015image}. In their research, a dataset of scene graphs of images was human-annotated manually. An arbitrary graph was made and used as a query to retrieve equivalent images having a similar graph. A Conditional Random Field model 
and a maximum posterior technique was employed to measure the similarity of graphs. Many algorithms were invented to generate scene graphs from images
\cite{yang2018graph,motifnet,tang2019learning,FNet}
since this structure and the dataset about it was public. By virtue of them, some studies about image-text retrieval \cite{wang2020cross,shi2019knowledge} have followed the scene graph approach recently. Shi et al. \cite{shi2019knowledge} created a scene concept graph based on a popular scene graph dataset \cite{krishna2017visual} and used it to expand the detected concepts in images to extract visual features. Cosine similarity was used to measure the similarity between these features and the semantic features produced by applying an RNN module to captions and rank them to perform the retrieval. The scene graph matching (SGM) model from Wang et al. \cite{wang2020cross} takes a scene graph from both images and captions as input then extracts the graph features by using their own designed encoders called visual scene graph and textual scene graph for each modality. Then, one can promptly calculate the similarity between the two graphs based on each part of the first graph's agreement with every part of the second graph. Despite getting state-of-the-art results among the scene graph approaches, the SGM could still be enhanced. Firstly, its similarity calculating process only takes the small parts of a graph into account, hence neglect the overall detail. Secondly, the SGM model does not apply any normalization technique, resulting in easy overfitting. Besides, using ResNet \cite{he2016deep} as the backbone of the structure also limits the model's potential performance. There have been several recent CNN-based networks that are more accurate than the ResNet.

Recently, some researchers have applied the attention technique to both visual and textual data to learn the interaction between the components themselves, then combine with another attention module to fuse two modalities together \cite{chen2020imram,wei2020multi}. Chen et al. \cite{chen2020imram} presented a novel structure that iteratively finds matching compartments between images and sentences and refined them progressively. Meanwhile, the multi-modality cross attention network employs the compelling Transformer \cite{vaswani2017attention} technique on images and semantic data to get the fine-grained relationship information in detail. They both manage to get the best results so far in the domain. Nevertheless, none of them use scene graphs in their approach, which is our main focus of this research.

In this paper, we try to improve the performance of the SGM method by employing a graph convolution model to capture the overall information of a graph which can be considered as one of its weaknesses mentioned above. Our model utilizes scene graph structures for images and sentences and encodes them into vector forms for storing overall essential information to calculate the similarity between graphs instead of operating at the graph level exclusively. We name this model ``Local and Global Scene Graph Matching'' (LGSGM). 

Our primary contributions are threefold. Firstly, the SGM measured the similarity between graphs by dividing them into sub-graphs, hence diminishing the overall detail of a graph. We overcome this problem by building a graph embedding module to summarise all information of a graph into a vector form and combine it with the graphical form to calculate the similarity of a pair of graphs. Secondly, the efficient network \cite{tan2019efficientnet} is applied to extract the visual features of detected objects in images instead of using conventional Residual structure \cite{he2016deep} in the SGM model as the former network is shown to be more accurate although having a simpler design. Moreover, we also integrate normalization techniques including Dropout and Batch normalization 
to mitigate the potential of overfitting. Thirdly, we run an evaluation on the Flickr30k dataset for the image-text matching problem to facilitate a comparison with the SGM model \cite{wang2020cross} as the baseline, which also uses scene graphs to support retrieval. We choose the SGM model as the baseline due to its state-of-the-art result in this field for models that employ scene graphs in the structure.

\section{Methodology}
\begin{figure*}[ht!]
  \centering
  \includegraphics[width=0.9\textwidth]{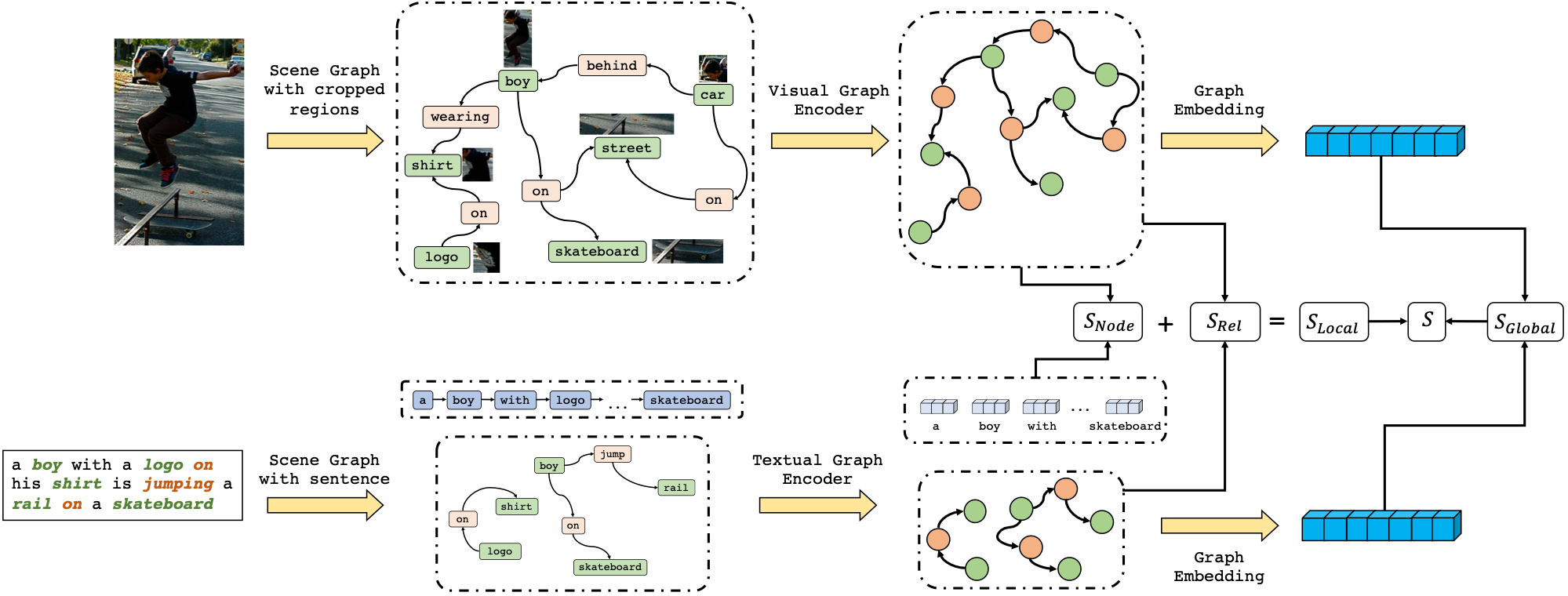}
  \caption{Our proposed LGSGM pipeline follows the SGM baseline. A scene graph of an image is firstly extracted at the preprocessing stage. The scene graph is then encoded to get the visual feature graph used to compute the local similarity score. A text also goes through a similar process to get the textual feature graph. Both feature graphs are embedded into vectors by a shared graph embedding model. The global similarity is then calculated based on their vector-level forms and combine with the local score to get the final similarity.}
  \label{fig:workflow}
\end{figure*}

This section will describe how our LGSGM model approaches the retrieval challenge as a graph similarity ranking problem which is inspired by the SGM. Figure \ref{fig:workflow} depicts the workflow of LGSGM that integrates an extra graph embedding stage to the SGM baseline. Initially, our preprocessing data stage starts with the scene graph construction. The graphs of all images and texts are extracted beforehand and stored in the database. A generated scene graph of a query, which can be an image or a sentence, then goes through a graph encoder module to get a graph feature.
Meanwhile, each sample in the remaining modality database also follows the same scheme but separately. All feature graphs now are passed through a Siamese-structured \cite{koch2015siamese} graph embedding layer to learn the summarised attributes of them, and the vector-level features of those graphs are obtained. The similarity of a pair of graphs, which is a query graph and a sample graph in the database, now can be calculated based on the feature graphs and their embedded vectors. Finally, these scores are ranked descending to find the most relevant answers from the database.

\subsection{Visual Graph Encoder}
\begin{figure}[h]
  \centering
  \includegraphics[width=0.69\linewidth]{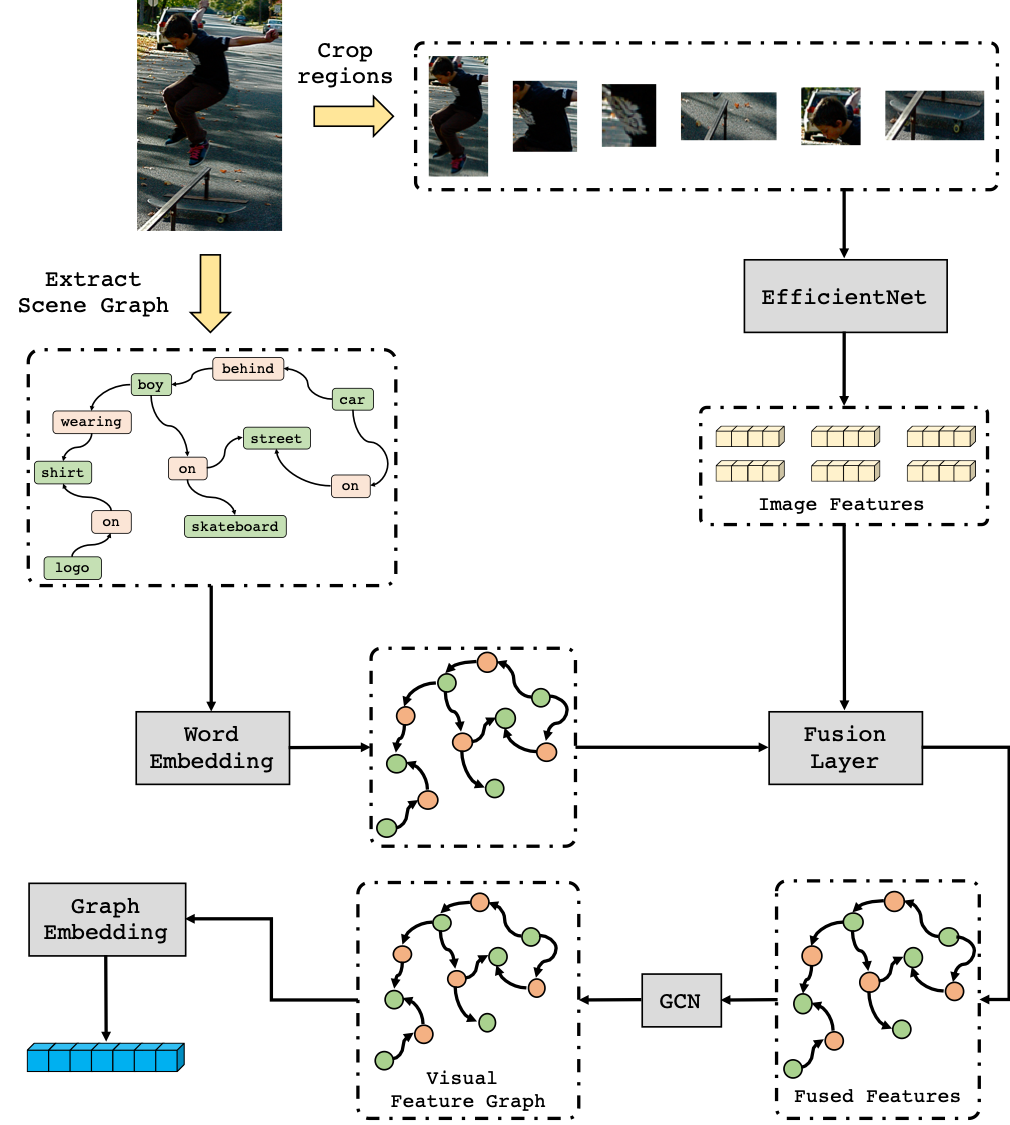}
  \caption{This figure presents the workflow of an image in our proposed model. First, regions of detected objects and a scene graph of the image are extracted. The regions are fed into a CNN-based model, which is the EfficientNet \cite{tan2019efficientnet} in our setting, to get the image features while the scene graph goes through the word embedding layer to get the semantic features of its nodes and edges. These features are updated by combining with corresponding image features by a fusion layer. A convolutional graph network is applied to the graph, and the visual feature graph is obtained. The graph is finally fed to the Siamese graph embedding model to get its global representation.}
  \label{fig:visual_branch}
\end{figure}

A scene graph of an image can be established beforehand by using any available scene graph generation method. 
The graph $G$ after extracted can be seen as a tuple $G = (O, B, R)$ where: 
\begin{itemize}
    \item $O = \{o_1, ..., o_{N_o}\}$ is the set of $N_o$ semantic labels of detected objects appearing the image.
    \item $B = \{b_{o_1}, ..., b_{o_{N_o}} | b_{o_i} \in \mathbb{R}^4\}$ is the set of bounding boxes where $b_{o_i}$ is the coordinate of the box of the corresponding object $o_i$ in the set $O$.
    \item $R = \{r_1, ..., r_{N_r}\}$ is the set of $N_r$ predicted relations between objects in the image. Each $r_m$ is a tuple of $(o_i, p_{ij}, o_j)$ where $p_{ij}$ is the label of the association between two objects $o_i$ and $o_j$. It is noted that $R$ can be seen as the set of edges connecting nodes, which are objects in $O$, of the scene graph $G$.
\end{itemize}
As shown in Figure \ref{fig:visual_branch}, after the graph is constructed, we embed the label of the nodes and edges into vectors with the help of trainable word embedding layers to gain their semantic information. Besides, there is also rich information from images themselves. We extract an image feature for each object in $O$ based on its associated bounding box in $B$. Regarding an edge $p_{ij}$, its image feature can be computed through the union regions of two boxes $b_{o_i}$ and $b_{o_j}$. The feature of one node or an edge now is the fusion of its image feature and semantic features. We update these features to learn the connection between nodes and edges features by using a convolutional graph neural network \cite{wu2020comprehensive} to get the final visual feature graph.

\textbf{Word Embedding}. We exploit the information of semantic labels by embedding names of objects and relations in $O$ and $R$ into vectors with two distinct embedding layers for each. Each object label $o_i$ and predicate label $p_{ij}$ is transformed into one-hot vector $I_{o_i}$ and $I_{p_{ij}}$.
Their embedding vectors, $e_{o_i}$ and $e_{p_{ij}}$, can be calculated as
\begin{equation}
e_{o_i}=W_oI_{o_i} \text{ ,  } W_o \in \mathbb{R}^{d_{W}\times\:C_o}
\end{equation}
\begin{equation}
e_{p_{ij}}=W_pI_{p_{ij}} \text{ ,  } W_p \in \mathbb{R}^{d_{W} \times\: C_p}
\end{equation}
where $W_o$ and $W_p$ are the trainable parameters in the layer, $C_o$ and $C_p$ is the number of categories of objects and relations supported in the scene graph generation method.
We set $d_{W}=300$ and utilise the pretrained embedding Glove model 
to initialise $W_o$ and $W_p$. 

\textbf{Image Features}. An image feature $v_{o_i}$ of an object $o_i$ can be obtained by applying a pretrained CNN-based network to the cropped region $b_{o_i}$ around the object. Similarly, the cropped region for an edge $p_{ij}$, which is the area covering both $b_{o_i}$ and $b_{o_j}$ regions, also be used to extract its $d_{I}$-dimension image feature $v_{ij}$.

\textbf{Fused Features}. A simple trainable, fully connected neural network is applied to the semantic feature's concatenated vector and the image feature to get the fused feature that can combine both modality detail. The fused representation of a node and an edge is achieved as followed:
\begin{equation}
u_{o_i} = f_{act}(W_u[v_{o_i}\:,\:e_{o_i}])
\label{eq:u_o}
\end{equation}
\begin{equation}
u_{ij} = f_{act}(W_u[v_{ij}\:,\:e_{ij}]),
\label{eq:u_p}
\end{equation}
where $[.]$ is the concatenating operation, $f_{act}$ is an activation function, and $W_u \in \mathbb{R}^{d_{F} \times (d_I \: + \: d_W)}$ is the trainable parameters.

\textbf{Graph Network}. We use a Convolutional Graph Network (GCN) to learn the connection and update the fused features of nodes ($u_{o_i}$) and edges ($u_{ij}$) of the graph. The GCN works similarly to a normal CNN operation but more flexible to many types of graph-structured data than only the grid data of a conventional CNN. In this GCN model, the features of a node are only updated based on itself solely to mitigate noise effects from surrounding nodes \cite{wang2020cross}. In contrast, because an edge bridges two nodes together, its features should be related to those nodes' features to learn the association between them. Regarding an n-layer GCN model, updated features of a node and an edge at the $l^{th}$ layer can be formulated as follows:
\begin{equation}
h^l_{o_i} = \text{MLP}_o(h^{l-1}_{o_i})
\label{eq:gcn_o}
\end{equation}
\begin{equation}
h^l_{p_{ij}} = \text{MLP}_p([h^{l-1}_{o_i}\:,\:h^{l-1}_{p_{ij}}\:,\:h^{l-1}_{o_j}]),
\label{eq:gcn_p}
\end{equation}
where $\text{MLP}_o$ and $\text{MLP}_p$ are two separate neural network models, and $h^0_{o_i}=u_{o_i}$ and $h^0_{p_{ij}}=u_{ij}$. The final output of the visual feature graph is the updated features of nodes and edges which are denoted as $h_{o_i}$ and $h_{p_{ij}}$.

\subsection{Textual Graph Encoder}
\begin{figure*}[h]
  \centering
  \includegraphics[width=0.9\linewidth]{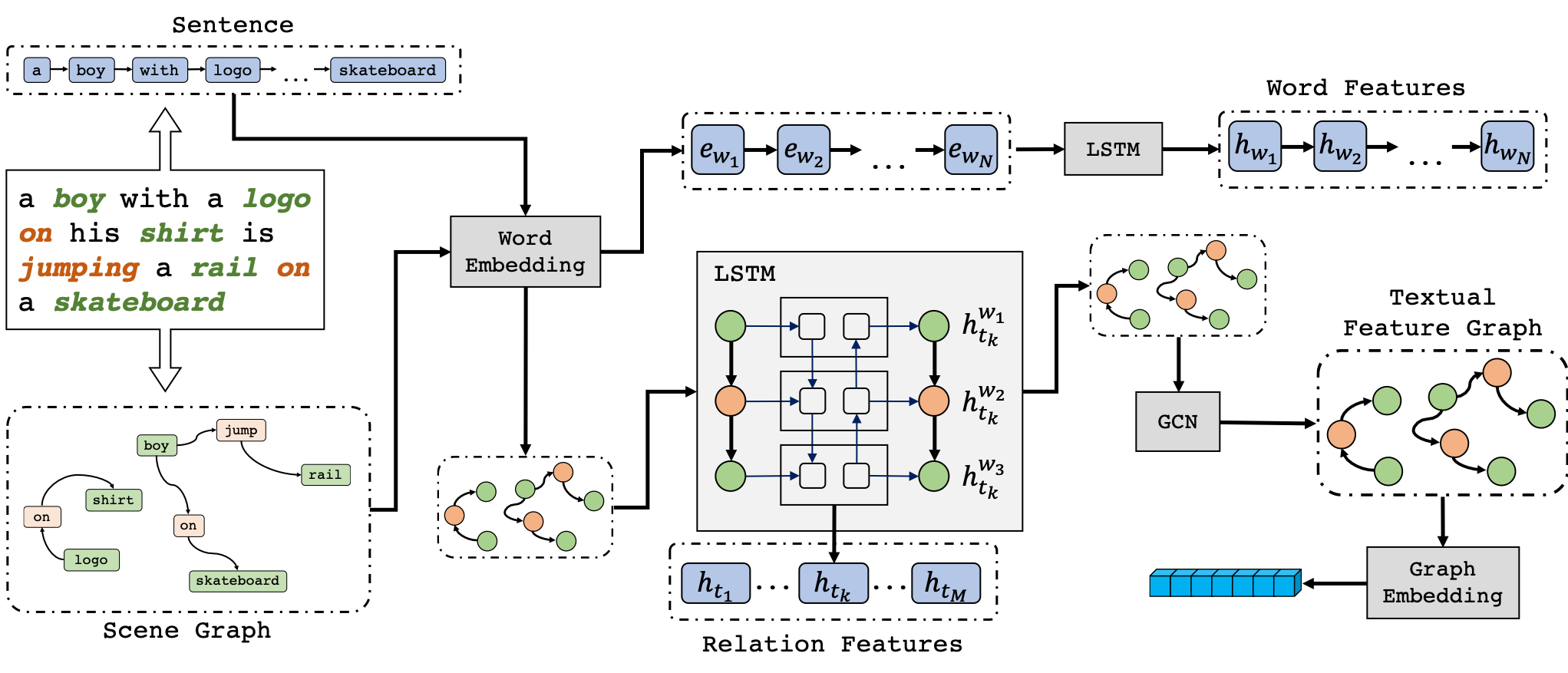}
  \caption{This figure presents the workflow of a sentence in our proposed model. Both scene graphs and each word in the sentence go through a word embedding to learn the semantic features. Two distinct LSTM models are applied to the sentence and the triplet relations in the graph to get the word and the relation features used to compare with the visual feature scene graph to get the local similarity. Each encoded node and edge in the graph after the LSTM model then form into a graph with will be fed into a graph convolutional network to update their features and create the textual feature graph. Finally, this graph is embedded into a vector compared with that of visual data to measure the global similarity score. The $N$ and $M$ in the figure indicate the number of words and number of relations in the sentence.}
  \label{fig:textual_branch}
\end{figure*}

Similarly, a sentence is also converted into a scene graph to describe the relationships in detail. These triplet information, such as \textit{"man-wears-hat"}, can be obtained by using SPICE technique \cite{anderson2016spice}. However, relying on the graph only will ignore other important clues in the sentence. Therefore two different modules are used to extract the data from both pathways as depicted in Figure \ref{fig:textual_branch}. Initially, every word in the sentence can be encoded by a word embedding module similar to that in the visual feature graph. Two LSTM models \cite{hochreiter1997long} are then employed to learn the features of each word in the sentence as well as the features of each extracted relation triplet. We expand the SGM model by introducing an additional GCN to the graph created from the triplets. This graph is used to summarise information of all relations in the sentence and compared to that of visual modality afterward.

\textbf{Word Embedding}. It is worth noting that there are a massive number of vocabularies in a real-world deployment. Hence, it is not suitable for the word embedding part to learn all words in the entire semantic modality database as an overfitting issue will occur in the test condition. We create the dictionary from the textual database (excluding the test set) and only learn some popular words. We replace the least common ones, whose appearance frequency less than 4, with the \textit{"unknown"} word, which is considered a popular vocabulary. During the test phase, words not included in the dictionary are converted to \textit{"unknown"}.

\textbf{Word Features}. After the embedding stage, the entire sentence goes to the bidirectional LSTM hierarchy to get the features from both forwards and backward ways. Then the representation of a word $w_i$ in the sentence, denoted as $h_{w_i}$, is the average of two paths and computed as follows:
\begin{equation}
    h_{w_i} = \frac{\overrightarrow{\text{LSTM}_w}(e_{w_i}, \overrightarrow{h_{w_{i-1}}})
    + \overleftarrow{\text{LSTM}_w}(e_{w_i}, \overleftarrow{h_{w_{i+1}}})}{2},
\end{equation}
where $\text{LSTM}_w$ is the bi-LSTM model, $e_{w_i}$ is the word embedding vector whilst $\overrightarrow{h_{w_{i-1}}}$ and $\overleftarrow{h_{w_{i+1}}}$ are the hidden states of the word $w_i$ from forward and backward directions.

\textbf{Graph Network}. We organise the set of relations of a caption as $T = \{t_1, ..., t_{N_t}\}$ with $N_t$ is the total number of relations in the sentence parsed by SPICE. Each triplet $t_m$ is a sequence of words $(w_i, w_{ij}, w_j)$ where the word $w_{ij}$ is the association between two objects $w_i$ and $w_j$ such as ("man", "wears", "hat"). We use another bidirectional LSTM structure, named $\text{LSTM}_t$, and share it among all triplet $t_i$ for $i \in [1, N_t]$. The feature of a word in a triplet is the hidden state itself while the feature of entire triplet is the last state of the sequence model. Specifically, they are calculated as: 
\begin{equation}
    h^{w_n}_{t_k} = \frac{\overrightarrow{\text{LSTM}_t}(e^{w_n}_{t_k}, \overrightarrow{h^{w_{n-1}}_{t_k}})
    + \overleftarrow{\text{LSTM}_t}(e^{w_n}_{t_k}, \overleftarrow{h^{w_{n+1}}_{t_k}})}{2},
\end{equation}
\begin{equation}
    h_{t_k} = \frac{\overrightarrow{h^{w_{i}}_{t_k}}+\overleftarrow{h^{w_{j}}_{t_k}}}{2},
\end{equation}
where triplet $t_k$ is a sequence of words started with $w_i$ and ended with $w_j$, $e^{w_n}_{t_k}$ is the word embedding vector of $w_n$ in $t_k$. An extra GCN model having a similar structure with that in visual modality (Eq. \ref{eq:gcn_o} and Eq. \ref{eq:gcn_p}) is then applied to the graph formed by the set $T$. Particularly, the graph will have $h^{w_n}_{t_k}$ as feature of nodes and $h^{w_{ij}}_{t_k}$ for edges which are all go through two neural networks to get the graph-level features of nodes and edges which are denoted as $hg^{w_n}_{t_k}$ and $hg^{w_{ij}}_{t_k}$.

\subsection{Graph Embedding}
After feature graphs of both modalities are extracted, we embed them into vector forms using an attention mechanism called multi-scale node attention \cite{bai2019unsupervised}. Given a graph $G$ having $N$ nodes $\{h_1, ..., h_N\}$ and $M$ edges $\{r_1,...,r_M\}$ where $h_i$ and $r_j \in \mathbb{R}^d$ is a feature of a node and an edge accordingly, the embedded vector of $G$, noted as $a_G \in \mathbb{R}^{2d}$, is obtained by the following formula:
\begin{eqnarray}
    a^{h}_G &=& \sum_{n=1}^{N} \sigma (h^\text{T}_n\text{ReLU}(W_h(\frac{1}{N}\sum_{m=1}^{N}h_m)))h_n,\\
    a^{r}_G &=& \sum_{n=1}^{M} \sigma (r^\text{T}_n\text{ReLU}(W_r(\frac{1}{M}\sum_{m=1}^{M}r_m)))r_n,\\
    a_G &=& [a^{h}_G\:, a^{r}_G]
\end{eqnarray}
where $\sigma$ and ReLU indicate the sigmoid and rectified linear unit activate function, $W_h$ and $W_r \in \mathbb{R}^{d \times d}$ is the training parameter in the model, subscript T denotes the transpose operation, and $[.]$ is the concatenation. We share this attention structure to the visual feature graph and textual feature graph as a siamese-model to get their vector-level representation $a_V$ and $a_T$ respectively. It is noted that the node features and edge features of a visual graph are $h_{o_i}$ and $h_{p_{ij}}$ while those of textual graph are $hg^{w_n}_{t_k}$ and $hg^{w_{ij}}_{t_k}$.

\subsection{Similarity Function}
The similarity between two scene graphs is measured using both local and global approaches. In the local form, each part of a graph is compared to every part of other graphs; for instance, matching on each node-level feature and on each edge-level feature. Regarding the global approach, the embedding vectors of the entire two graphs are used. This similarity score can be used to rank the retrieved result as higher means more relevant answers.

\textbf{Local similarity}. The local score is the sum of a node score and an edge score. The former score is the average of the matching scores of all words in a sentence with their most relevant object detected in an image. The relevant score of a word $w_i$ in the caption with an object in the image $o_k$ is their dot product $h_{w_i}^\text{T}h_{o_k}$. Assuming there are $N_w$ words in the caption and $N_o$ objects in the image, the node score is calculated as:
\begin{equation}
    S_{Node} = \frac{1}{N_w} \left(\sum_{i=1}^{N_w}\underset{k\in[1,N_o]}{\mathrm{max}}h_{w_i}^\text{T}h_{o_k}\right)
\end{equation}

Similarly, the edge score is the mean score between relations in the textual data with their most matching edge in the visual graph and is formulated as:
\begin{equation}
    S_{Rel} = \frac{1}{N_t} \left(\sum_{i=1}^{N_t}\underset{p_{ij}\in R}{\mathrm{max}}h_{t_i}^\text{T}h_{p_{ij}}\right)
\end{equation}

The local similarity then can be obtained as follows:
\begin{equation}
    S_{Local} = S_{Node} + S_{Rel}
\end{equation}

\textbf{Global similarity}. The global vectors provide the overall information of graphs. We use conventional cosine distance to measure the degree of matching between two entire graphs. Specifically, the similarity score, $S_{Global}$, of two embedding vectors of visual graph $a_V$ and textual graph $a_T$ is:
\begin{equation}
    S_{Global} = \frac{a^\text{T}_Va_T}{\|a_V\| \: \|a_T\|}
\end{equation}
Finally, the similarity between two graphs takes both local and global details into account, hence is calculated by the sum of them, which is 
\begin{equation}
    S = S_{Local} + S_{Global}
\end{equation}

This bi-level similarity score is also used in the loss function during the training phase where we apply the hardest negative triplet loss as in \cite{faghri2017vse++,wang2020cross}. The loss function is:
\begin{equation}
    L(k,l) = \text{max}(0, m - S_{kl} + S_{k\hat{l}}) + \text{max}(0, m - S_{kl} + S_{\hat{k}l}),
\label{eq:loss}
\end{equation}
where $m$ is a margin hyperparameter, $S_{kl}$ is the similarity score of a true pair of image $k$ and its caption $l$, and $\hat{k}$, $\hat{l}$ is the least matching image and sentence with $l$ and $k$ in the mini-batch, respectively.

\section{Dataset and Metrics}
We evaluate our proposed model on the Flickr30k dataset \cite{flickr30k}, which is one of the most popular datasets in this image-text matching field due to its high quality of textual annotation compared to others \cite{plummer2015flickr30k}. The dataset consists of $31,783$ images, and each of them has five corresponding captions. An example of image-sentence pairs and their generated scene graph can be illustrated in Figure \ref{fig:scenegraph}. We split the Flickr30k data into training, validating, and testing sets, so that the number of the training images is $29,783$, whilst the two latter subsets both have $1,000$ images.

The evaluation metric we choose is a common Recall at K (R@K) value that has been used in many kinds of research \cite{wang2020cross,wang2019position,chen2020imram,wei2020multi}. The R@K is the proportion of queries that we find their correct matching answers in the top K of the ranking result. In our experiments, we evaluate three values of K, which are 1, 5, and 10.

\section{Experiments and Results}
In this section, we run our proposed model on the Flickr30k dataset then compare the baseline that is chosen as the SGM model as both our LGSGM and the SGM have a similar scene graph approach.

\subsection{Model Setting}
At the scene graph generation stage, we use Neural Motifs \cite{motifnet} 
to get the similar graphs as described in \cite{wang2020cross}. We only keep the top $N_o=36$ detected objects and $N_r=25$ predicates in the result sorted by their confidence scores for each scene graph. Regarding visual features of objects, we employ the EfficientNet, which is considered to be more accurate in the image classification challenge, although fewer parameters \cite{tan2019efficientnet} than ResNet \cite{he2016deep} as used in the baseline. The region surrounding the object is firstly cropped based on its bounding box and resized to the desired format, then goes through the net, which is excluded from the classification layer, to get the feature. We set $d_I=d_F=2048$ and choose the EfficientNet-b5 model pretrained on ImageNet 
to get the 2048-dim feature vectors as comparable with the baseline. The number of layers in GCN models for both modalities and the LSTM models is $1$. The dimension of the output vector from GCN is configured to be $d=1024$, which is also the size of the hidden state of LSTM models. We also apply Dropout 
and Batch-normalisation 
to avoid the potential overfitting in both hierarchies. We use Swish \cite{ramachandran2017searching} as an activate function when creating fused features (Eq. \ref{eq:u_o} and Eq. \ref{eq:u_p}) and in GCN models. We also tried other activation functions (such as ReLU, Leaky ReLU, or Tanh), but Swish has the best performance among these functions. The margin $m$ in the loss function (Eq. \ref{eq:loss}) is selected as $0.35$. We train the model with the batch size of $128$ at the learning rate of $0.0003$ with Adam optimizer. 

\subsection{Comparison with Baseline and other Methods}
Although our primary focus is to compare with the baseline SGM, we also show the result of other state-of-the-art techniques to provide a comprehensive perspective on the performance of our proposed approach. Besides the SGM, other competing models are:
\begin{itemize}
    \item PFAN \cite{wang2019position} which uses the position focused attention network to extract the features of the location of objects in images.
    \item IMRAM \cite{chen2020imram} with the attention mechanism to learn the matching fragments between images and sentences.
    \item MMCA \cite{wei2020multi} applying attention and transformer compartment to exploit the relationship between objects in images and words in texts within themselves.
    \item GSMN \cite{liu2020graph} that is also a graph-based approach but connects all detected objects within an image to create a graph then compares to that of a text.
\end{itemize}

The LGSGM and others' results on Flickr30k can be depicted in Table \ref{tab:flickr30k}. The value in bold is the highest number in that metric. Caption retrieval indicates that a model needs to find texts that are relevant to a query image. On the contrary, image retrieval is used when a query is a sentence. It is important to note that other models' metrics are taken from their original report since we use the same subset for training, validating, and testing.

\begin{table*}[ht!]
  \caption{Performance of models on Flickr30k Dataset. R-Sum is the sum of all recall metrics.}
  \label{tab:flickr30k}
  \begin{tabular}{|c|c|c|c|c|c|c|c|c|}
    \hline
    \multirow{2}{*}{\textbf{Methods}} & \multicolumn{4}{|c|}{\textbf{Caption retrieval}} & \multicolumn{4}{|c|}{\textbf{Image retrieval}}\\
    \cline{2-9}
    & R@1 & R@5 & R@10 & R-Sum & R@1 & R@5 & R@10 & R-Sum\\
    \hline
    PFAN \cite{wang2019position} & 70 & 91.8 & 95.0 & 256.8 & 50.4 & 78.7 & 86.1 & 215.2 \\
    \hline
    IMRAM \cite{chen2020imram} & 74.1 & 93.0 & 96.6 & 263.7 & 53.9 & 79.4 & 87.2 & 220.5 \\
    \hline
    MMCA \cite{wei2020multi} & 74.2 & 92.8 & 96.4 & 263.4 & 54.8 & 81.4 & 87.8 & 224.0 \\
    \hline
    GSMN \cite{liu2020graph} & \textbf{76.4} & \textbf{94.3} & \textbf{97.3} & \textbf{268.0} & \textbf{57.4} & 82.3 & 89.0 & 228.7\\
    \hline
    SGM \cite{wang2020cross} & 71.8 & 91.7 & 95.5 & 259.0 & 53.5 & 79.6 & 86.5 & 219.6\\
    \hline
    LGSGM (Ours) & 71 & 91.9 & 96.1 & 259.0 & \textbf{57.4} & \textbf{84.1} & \textbf{90.2} & \textbf{231.7}\\
    \hline
\end{tabular}
\end{table*}

It is easier for all models to find captions when an image is given as a query than the reversed retrieving because the caption retrieval section's scores are higher than those in image retrieval. Regarding the caption retrieval, the GSMN using an ensemble setting performs best among all models, where the differences in the R-Sum can be up to more than $10\%$. This model is more accurate than ours with $3\%$ on average of recall. Both attention-based techniques, IMRAM and MMCA, share similar results as there is no significant gap in all recall metrics. It is also true for SGM and LGSGM models. Although we obtain higher scores at R@5 and R@10 with a margin less than $0.5\%$, the SGM model has a slightly better R@1 metric than ours, which are $71.8\%$ and $71\%$, respectively. The PFAN model performs worst with the lowest R@1 and R@10. However, our LGSGM achieves the highest scores on all three recall metrics in the image retrieval field, which are $57.4\%$, $84.1\%$, and $90.2\%$ accordingly. It makes the proposed method have the best R-Sum of $231.7\%$, which is $3\%$ higher than the GSMN in the second place. In specific, our R@5 and R@10 are better than that of GSMN roughly by $1.8\%$ and $1.2\%$ respectively. Compared with the SGM, our improvement creates a huge increase of $12.1\%$ in total recall. The PFAN is still the model with the lowest recall, while the R-Sum of MMCA is $4.5\%$ higher than IMRAM.

In general, the proposed LGSGM surpasses the SGM model by a large margin, which is also our main contribution. Our model and GSMN, which is also a graph-based method, are the top-2 methods in the experiment, showing the usefulness of the graph structure in this field. Nevertheless, GSMN achieves higher recall than ours with $496.7\%$ compared to $490.7\%$ of our model. It might be due to its dense graph structure where this model connects all of the objects in an image while the scene graph structure only captures some detected relations between them. Although the attention approaches are better than our scene graph model in the image-to-text retrieval, our network still manages to get the highest score on the remaining experiment. Moreover, LGSGM manages to score better than those models concerning the sum of recall of both image retrieval and caption retrieval experiments. With our state-of-the-art result in the text-to-image retrieval section, it opens a wide range of applications of our structure in the field of finding images that are relevant to the given description. For instance, one potential application is in lifelogging retrieval, where a graph-based method has shown its promising performance \cite{mmm2021graph}.

\section{Conclusion}
In this research, we address an issue that remained in the state-of-the-art scene graph matching model in which the global detail of graphs is ignored during the graph encoding phrase. We propose a graph embedding module that can address that concern by summarising the overall information of a graph into a vector form. Our LGSGM method, therefore, can measure the similarity between images and captions based on their input scene graphs with both local and global views. Using a lighter and more accurate EfficientNet to extract features combining with normalizing techniques to mitigate the over-fitting problem, our model can surpass the baseline and achieve the new state-of-the-art results for those using scene graph as input in image-text retrieval challenge.

\section{Acknowledgments}
This publication has emanated from research supported in party by research grants from Science Foundation Ireland 
under grant numbers SFI/12/RC/2289, SFI/13/RC/2106, and 18/CRT/6223. 

\bibliographystyle{unsrt}
\bibliography{ref.bib}

\end{document}